# BLIND NORMALIZATION OF SPEECH FROM DIFFERENT CHANNELS AND SPEAKERS

David N. Levin, Department of Radiology, University of Chicago (d-levin@uchicago.edu)

## ABSTRACT

This paper describes representations of time-dependent signals that are invariant under any invertible time-independent transformation of the signal time series. Such a representation is created by rescaling the signal in a non-linear dynamic manner that is determined by recently encountered signal levels. This technique may make it possible to normalize signals that are related by channel-dependent and speaker-dependent transformations, *without* having to characterize the form of the signal transformations, which remain unknown. The technique is illustrated by applying it to the time-dependent spectra of speech that has been filtered to simulate the effects of different channels. The experimental results show that the rescaled speech representations are largely normalized (i.e., channel-independent), despite the channel-dependence of the raw (unrescaled) speech.

## 1. INTRODUCTION

Humans perceive the information content of ordinary speech to be remarkably invariant, even though the signal may be transformed by significant alterations of the speaker's voice, the listener's auditory apparatus, and the channel between them. Yet there is no evidence that the speaker and listener exchange calibration data in order to characterize and compensate for these distortions. Evidently, the speech signal is redundant in the sense that listeners extract the same content from multiple acoustic signals that are transformed versions of one another. In earlier reports [1-4], the author showed how to design sensory devices that behave in this way. In such devices, the signal is represented by a non-linear function of its instantaneous level at each time, with the form of this scale function being determined by recently encountered signal levels. This dynamically rescaled signal is invariant if the original signal levels are transformed by any invertible time-independent function. This is because the transformation's effect on the signal level at any time is cancelled by its effect on the scale function at that time. The method was illustrated by applying it to analytic examples, simulated signals, acoustic waveforms of human speech, time-dependent spectra of bird songs, and time-dependent spectra of synthetic speech-like signals [1-4]. In this paper, we apply the method to time-dependent spectra of natural speech that has been filtered in order to simulate the effects of different channels.

## 2. THEORY

Let $x(t)$ ( $x_k(t)$, $k = 1,2,...,N$ ) denote $N$ parameters (e.g., cepstral coefficients) that characterize the short-term Fourier spectrum of speech at time $t$. In the following, we show how the speech trajectory $x(t)$ defines a differential geometry that determines a special coordinate system $s(x)$ on the $x$ manifold. Speech is invariantly represented in this coordinate system in the following sense: if the $x$ trajectory is subjected to any invertible transformation, the representation of the transformed trajectory in *its s* coordinate system is the same as the representation of the untransformed speech in *its s* coordinate system.

Consider a point $y$ in a region of the $x$ manifold that is densely sampled by the speech trajectory. Define $g^{kl}$ to be the average outer product of the time derivatives of the speech trajectory as it passes through a small neighborhood of $y$: $g^{kl} = \left\langle \frac{dx_k}{dt} \frac{dx_l}{dt} \right\rangle_{x(t) \sim y}$. As long as this neighborhood contains $N$ linearly independent time derivatives, $g^{kl}$ is positive definite, and its inverse $g_{kl}$ is well defined and positive definite. Under any transformation $x \rightarrow x' = x'(x)$, $\frac{dx}{dt}$ transforms as a contravariant vector. Therefore, $g^{kl}$ and $g_{kl}$ transform as a contravariant and covariant tensors, respectively. This means that $g_{kl}$ can be taken to define a metric on the manifold, and a process for parallel transporting vectors across the manifold can be derived from this metric by means of the methods of Riemannian geometry. For example, the parallel transport process can be defined by requiring that it not change the lengths of vectors with respect to the metric. Now suppose that $N$ linearly-independent "reference" vectors $h_a$ ( $a = 1,2,...,N$ ) can be defined at a special "reference" point $x_0$ on the manifold. For example, this reference information may be specified by prior knowledge, analogous to the way a choir leader gives the singers prior knowledge of the origin of the musical scale by playing a pitch pipe before the concert. Alternatively, it may be possible to derive the reference information from the intrinsic properties of the speech trajectory itself [1, 4]. For example, $x_0$ could be chosen to be an extremum of the manifold's curvature scalar. The reference vectors can be parallel transported across the manifold to determine the $s$ coordinates of any point $x$. For example, the point $x$ can be assigned the coordinates $s$ ( $s_k$, $k = 1,2,...,N$ ), if it is reached by starting at the reference point and by parallel transporting $h_1$ along itself $s_1$ times, then parallel transporting $h_2$ along itself $s_2$ times, …, and finally parallel transporting $h_N$ along itself $s_N$ times. Notice that this parallel transport process is independent of the nature of manifold's $x$ coordinate system.

Therefore, as long as the reference point/vectors can be identified in a coordinate-independent manner, the *s* representation of the speech trajectory will also be coordinate-independent. Because any invertible transformation of the speech trajectory is equivalent to a change of the manifold's *x* coordinate system, the *s* representation of a speech trajectory is unaffected by any invertible transformation of the data. This means that the *s* representations of the speech trajectories from two different channels or speakers will be identical, as long as the two unrescaled trajectories differ by an invertible transformation. Likewise, the *s* representations of two parameterizations of a given speech sample (e.g., cepstral coefficients vs. spectral coefficients vs. mel cepstral coefficients vs. …) will be the same, as long as the two parameterizations differ by an invertible transformation.

### 3. EXPERIMENTS WITH SPEECH SPECTRA

The experimental data were 27 seconds of speech (ten sentences) from a 32 year old Midwestern American male [5] that was digitized with 16 bits of precision at 8000 Hz. Short-term Fourier spectra were produced from the signals in a 16 ms Hamming window that was advanced in increments of 4 ms. Figure 1a shows the acoustic waveform and spectrogram of the speech signal corresponding to one of the sentences. The spectrum at each time point was parameterized by 53 cepstral coefficients, which were generated by the discrete cosine transformation (DCT) of the log of the spectral magnitude, after the spectral magnitude had been averaged in equally spaced 800 Hz bins. The cepstral coefficients of each of the 6853 short-term spectra defined a single point in a 53-dimensional space. Principal components analysis was used to reduce the dimensionality of these data. Figure 1b shows the trajectory $x(t)$ of the first two principal components during the course of the ten sentences. Notice that a fairly compact domain of the *x* manifold is sampled by the ten sentences. As outlined in Section 2, the speech trajectory was used to compute a metric on a *7 x 9* rectangular array of evenly spaced points that covered the most densely sampled region ($-29 \leq x_1 \leq 41, -24 \leq x_2 \leq 26$). A parallel transport process was derived from these metric samples by smoothly interpolating them and computing an affine connection from the resulting metric function. The reference point was chosen to be a particular point on the speech trajectory near the middle of the rectangular region on which the metric and affine connection were computed, and the reference vectors were chosen to be the time derivatives of the speech trajectory at two particular times when it passed near the reference point. The *s* coordinates of each point on the manifold were then derived by computing how many times $h_1$ and $h_2$ had to be parallel transported along themselves in order to reach that point. Figure 1c shows *s* isoclines for values of *s* between 5 (left), -10 (right), -5 (bottom), and 9 (top). Figures 1d and 1e show the *x* and *s* representations of a 292 ms segment of speech, corresponding to the word "ask" in Fig. 1a.

In order to simulate the effect of a different channel, the lowest and highest frequencies (0 - 500 Hz and 3500 - 4000 Hz) in each short-term Fourier spectrum were attenuated by Hamming filters with 500 Hz roll-offs. Figure 2a depicts the acoustic waveform and spectrogram of the signal in Figure 1a after it was filtered. Visual inspection shows the relative attenuation of the low and high frequencies of the filtered signal, which has a "hollow" sound [6]. Figure 2b depicts the trajectory of the first two principal components of the 53 cepstral coefficients of each spectrum, after it had been averaged in 800 Hz bins. Notice that the speech trajectory densely samples a relatively compact region that is not the same as that sampled by the unfiltered speech. The filtered speech trajectory was used to compute a metric on an *8 x 9* rectangular array of evenly spaced points that covered the most densely sampled region ($-44 \leq x_1 \leq 36, -21 \leq x_2 \leq 29$), and this metric was used to define a parallel transport process, as before. The reference point and vectors were chosen to be the location and time derivatives, respectively, of the filtered speech trajectory at the same time points as those used to derive reference information from the unfiltered trajectory. Finally, the reference vectors were subjected to parallel transport in order to derive an *s* coordinate system from the filtered trajectory. Figure 2c shows *s* isoclines for values of *s* between -11 (left), 7 (right), 6 (bottom), and -8 (top). Figures 2d and 2e show the *x* and *s* representations, respectively, of the filtered version of the word "ask", depicted in Figs. 1d and 1e. Notice that the *s* representations of the filtered and unfiltered sounds are similar, even though their raw (unrescaled) representations are quite different. The similarity of the *s* representations would probably be even more striking if we had retained a larger number of principal components of the cepstral coefficient data.

The unfiltered and filtered speech trajectories had similar *s* representations because those trajectories were related by an invertible transformation. Such a transformation existed because two conditions were satisfied: 1) an invertible transformation related each unfiltered spectrum's position in the space of spectral coefficients to the corresponding filtered spectrum's position in that space; 2) the spectral coefficients of the unfiltered and filtered speech were invertibly related to the corresponding cepstral coefficients. The first condition was satisfied because the filtering operation did not obliterate all of the differences between any pair of spectra in the sound; i.e., it did not map different spectra in the unfiltered speech onto the same filtered spectrum. This was because the filter's spectrum was non-zero for most frequencies, including frequencies at which pairs of unfiltered spectra differed. Because the above two conditions were satisfied in this example, there was an invertible transformation between the cepstral coefficients derived from unfiltered and filtered spectra at identical times. Therefore, to the extent that the higher principal components of the cepstral coefficients could be ignored, there was an invertible transformation between the trajectories in Figs. 1b and 2b. As shown in Section 2, the existence of this mapping guaranteed the filter-independence of the speech's $s$ representation. Note that, because the spectra were blurred prior to the cepstral transformation, the cepstral coefficients of the unfiltered and filtered spectra were related by a non-linear transformation, rather than a simple translation.

## 4. DISCUSSION

This paper describes a non-linear signal processing technique (called dynamic rescaling) for identifying the "part" of a signal that is invariant under any time-independent invertible transformation of the signal time series. This form of the signal (called its $s$ representation) is found by rescaling the signal at each time in a manner that is determined by its recent time course. Signals that are related by a time-*dependent* transformation (e.g., due to a time-dependent channel) may have different $s$ representations immediately after each change in the nature of the transformation. However, if the most recently encountered speech data are used to rescale the signal at each time point, the dynamic rescaling process adapts to the new form of the transformation, and the invariance of the signal's $s$ representation is re-established [1-4]. The length of this period of adaptation is $\Delta T$, the user-defined length of the signal history that is used to derive the dynamic scale at each time point (e.g, the 30-second speech history used in the experiment in Section 3). This kind of adaptive rescaling is illustrated elsewhere [2-3]

This technology may provide a useful "front end" for intelligent sensory devices, such as speech recognition and computer vision systems [7]. The signals from the system's detectors would be dynamically rescaled before they are passed to the system's pattern recognition module for higher level analysis. The rescaled representation of a stimulus is invariant under changes in observational conditions and changes in the stimulus itself that cause invertible transformations of the states of the system's detectors. Unlike conventional sensory systems, a device with this type of "representation engine" need not be periodically recalibrated with test stimuli, and its pattern recognition software need not be retrained to compensate for conditions causing signal transformations.

The experiment in Section 3 showed that the unfiltered and filtered utterances of any one speaker have the same dynamically rescaled representations (Figs. 1e and 2e). It can be argued that the utterances of two different speakers of the same message will produce spectral parameter trajectories having the same rescaled representations. This will be the case if: 1) there is an invertible transformation between the two speakers' vocal configurations when they utter the same message and 2) the detected spectral parameters (e.g., spectral or cepstral coefficients) are invertibly related to the corresponding speaker's vocal configurations. The first condition will be true if the two speakers mimic each other's vocal movements in a consistent manner. In other words, at each time point during the utterance when speaker #1 uses one of his/her vocal tract configurations, speaker #2 must consistently use one of his/her vocal tract configurations and vice versa. The second condition is likely to hold as long as the number of detected spectral parameters is more than twice as large as the number of degrees of freedom of the speakers' vocal tracts. This is because the embedding theorems of non-linear dynamics [8] state that almost every mapping from a $d$-dimensional space to a space of more than $2d$ dimensions is invertible. Because there is evidence [9] that the human vocal tract has 3-5 degrees of freedom, this condition will be satisfied if more than 6-10 spectral parameters are detected. Conditions 1-2 imply that there is an invertible transformation between the spectral parameter trajectories detected when the two speakers utter the same message. Therefore, it follows from Section 2 that these spectral parameter trajectories will have identical dynamically rescaled representations.

There is another interesting implication of the above-mentioned likelihood of an invertible relationship between a detected spectral parameter trajectory and the trajectory of the vocal apparatus configurations that produced it, Namely, the trajectory of parameterized speech must have the same dynamically rescaled representation as the trajectory of vocal movements during the utterance. Therefore, the rescaled trajectory of the detected signal is an "inner" property of the vocal tract motion. In fact, this is the reason *why* the rescaled detected signal is independent of "outer" features of the detection process (e.g., the nature of the channel and detected spectral parameters).

Humans tend to have similar perceptions of speech despite significant differences in their sensory organs and processing pathways. Furthermore, each individual has the remarkable ability to perceive the intrinsic constancy of speech even though its "sound" depends on the speaker and channel. These phenomena have been the subject of philosophical discussion since the time of Plato (e.g., his allegory of "The Cave"), and they have also intrigued modern neuroscientists. This paper shows how to design a sensory device that represents speech in a way that is largely independent of the nature of the speaker, channel, or detector. These representations are invariant because invertibly-related signals (e.g., signals from different speakers, channels, detectors) are the same signal viewed in different coordinate systems and because the rescaling process extracts the coordinate-independent relationship of that signal to the intrinsic differential geometric structure of the manifold. Perhaps, humans perceive speech in a nearly universal and constant manner because they too have the ability to appreciate its "inner" geometry. In any event, it will be interesting to see to what extent our different communication methods (e.g., different languages or types of music) have the same intrinsic geometry.

## 5. REFERENCES

[1] Levin, D. N., "Stimulus representations that are invariant under invertible transformations of sensor data", *Proc. of SPIE*, 4322: 1677-1688, 2001. To download, see http://www.geocities.com/dlevin2001/reprint1.html.

[2] Levin, D. N., "Universal communication among systems with heterogeneous 'voices' and 'ears' ", *Proc., Internat. Conf. on Advances in Infrastructure for Electronic Business, Science, and Education on the Internet*, Scuola Superiore G. Reiss Romoli S.p.A., L'Aquila, Italy, 6-12 August, 2001. See http://www. geocities.com/dlevin2001/reprint2.html.

[3] Levin, D. N., "Representations of sound that are insensitive to spectral filtering and parameterization procedures", *J. Acoust Soc. Amer.*, accepted for publication (in press), 2001. See http://www. geocities.com/dlevin2001/preprint1.html.

[4] Levin, D. N., "Sensor-independent stimulus representations", *Proc. Nat. Acad. Sci (USA),* accepted for publication (in press), 2001. See http://www. geocities.com/dlevin2001/preprint2.html.

[5] TIMIT corpus of the Linguistic Data Consortium, http://www.ldc.upenn.edu/.

[6] The sounds in this paper can be heard at: http://www. geocities.com/dlevin2001/reprint3.html.

[7] Levin, D. N., patents pending, 2000.

[8] Sauer, T., Yorke, J. A., and Casdagli, M., "Embedology", *J. Stat. Phys.*, 65: 579-616.

[9] Townshend, B., "Nonlinear prediction of speech signals". In: Casdagli, M. and Eubank, S. (Eds.), *Nonlinear Modeling and Forecasting*, Addison-Wesley, New York, 1992.

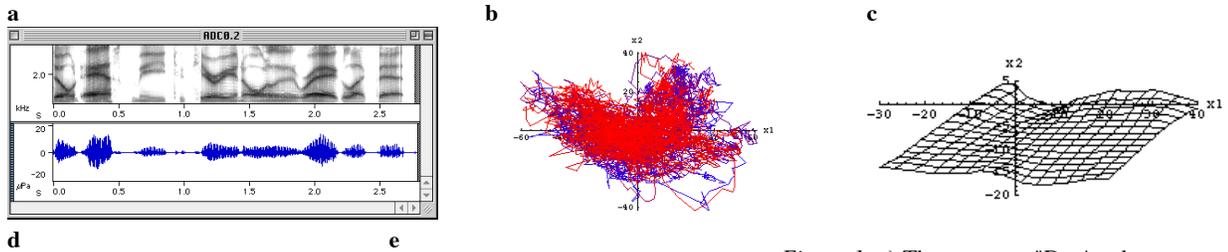

*Figure 1:* a) The sentence "Don't ask me to carry an oily rag like that". b) The $x$ trajectory of $a$ and nine other sentences. c) The $s$ scale derived from $b$; $s$ isoclines between 5 (left), -10 (right), -5 (bottom), and 9 (top). d) and e) The $x$ and $s$ trajectories, respectively, of the word "ask".

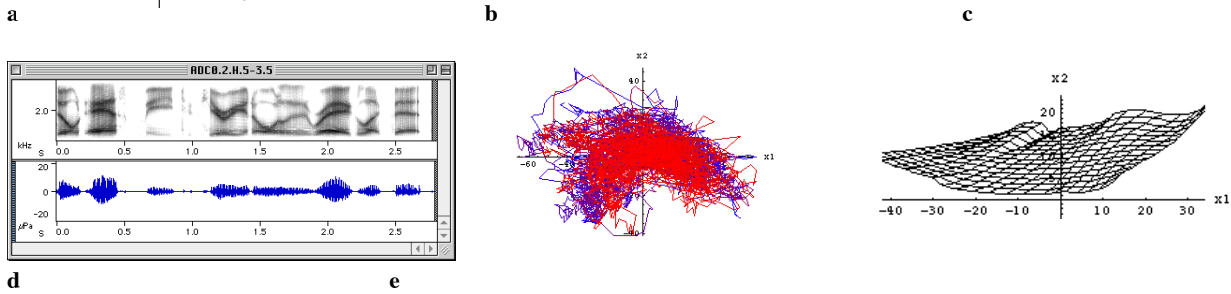

*Figure 2:* a) The filtered version of Fig. 1a. b) The $x$ trajectory of the filtered versions of the ten sentences in Fig. 1b. c) The $s$ scale derived from $b$; $s$ isoclines between -11 (left), 7 (right), 6 (bottom), and -8 (top). d) and e) The $x$ and $s$ trajectories, respectively, of the filtered word "ask".